\documentclass[letterpaper, 10 pt, conference]{ieeeconf}  

\IEEEoverridecommandlockouts
\usepackage{graphicx}
\usepackage{epstopdf}
\usepackage{amsmath}
\usepackage{amssymb}

\usepackage{subfigure}

\usepackage{multirow}
\usepackage{pbox}
\usepackage{algorithm}
\usepackage{algpseudocode}
\usepackage{titlesec}
\usepackage{bm}
\usepackage{url}
\usepackage{dblfloatfix} 
\usepackage{float}

\usepackage{algpseudocode,algorithm,algorithmicx}  
\algrenewcommand\algorithmicrequire{\textbf{Precondition:}}  
\algrenewcommand\algorithmicensure{\textbf{Postcondition:}}

\usepackage[utf8]{inputenc}
\usepackage{amsfonts}
\usepackage{amssymb}
\usepackage{mathtools}
\usepackage[normalem]{ulem}
\usepackage[nodisplayskipstretch]{setspace}

\usepackage{comment}
\usepackage[colorinlistoftodos,prependcaption,textwidth=1.5cm,textsize=tiny]{todonotes}

\usepackage[normalem]{ulem}

\newtheorem{problem}{Problem}
\newtheorem{lemma}{Lemma}

\newtheorem{corollary}{Corollary}[lemma]

\renewcommand{\baselinestretch}{0.95}

\usepackage{xcolor}

\usepackage[colorinlistoftodos,prependcaption,textwidth=1.5cm,textsize=tiny]{todonotes}

\title{Learning Enabled Fast Planning and Control in Dynamic Environments with Intermittent Information}
\author{Matthew Cleaveland$^\dagger$, Esen Yel$^\ddag$, Yiannis Kantaros$^\P$, Insup Lee$^\dagger$, Nicola Bezzo$^\mathsection$
\thanks{$^{\dagger}$Matthew Cleaveland and Insup Lee are with the Department of Computer Science and Precise Center, University of Pennsylvania, Philadelphia, PA 19104, USA. Email:{\texttt{ \{mcleav, lee\}@seas.upenn.edu}}}
\thanks{$^\ddag$Esen Yel was with the Department of Engineering Systems and Environment, University of Virginia, Charlottesville, VA 22904, USA. She is now with the Department of Aeronautics and Astronautics, Stanford University, Stanford, CA, 94305, USA. Email: {\texttt{esenyel@virginia.edu}}}
\thanks{$^\P$Yiannis Kantaros was with the Department of Computer Science and Precise Center, University of Pennsylvania, Philadelphia, PA 19104, USA. He is now with the Department of Electrical and Systems Engineering, Washington University of St. Louis, MO, 63130. Email: {\texttt{kantaros@seas.upenn.edu}}}
\thanks{$^{\mathsection}$Nicola Bezzo is with the Department of Electrical and Computer Engineering, and Link Lab, University of Virginia, Charlottesville, VA 22904, USA. Email: {\texttt{nb6be@virginia.edu}}}%
}
\date{February 2021}

\begin{document}

\maketitle

\begin{abstract}
This paper addresses a safe planning and control problem for mobile robots operating in communication- and sensor-limited dynamic environments. In this case the robots cannot sense the objects around them and must instead rely on intermittent, external information about the environment, as e.g., in underwater applications. The challenge in this case is that the robots must plan using only this stale data, while accounting for any noise in the data or uncertainty in the environment. To address this challenge we propose a compositional technique which leverages neural networks to quickly plan and control a robot through crowded and dynamic environments using only intermittent information. Specifically, our tool uses reachability analysis and potential fields to train a neural network that is capable of generating safe control actions. We demonstrate our technique both in simulation with an underwater vehicle crossing a crowded shipping channel and with real experiments with ground vehicles in communication- and sensor-limited environments.


\end{abstract}

\section{Introduction}
\label{ref:intro}



\begin{figure}
    \centering
    \includegraphics[width=0.48\textwidth]{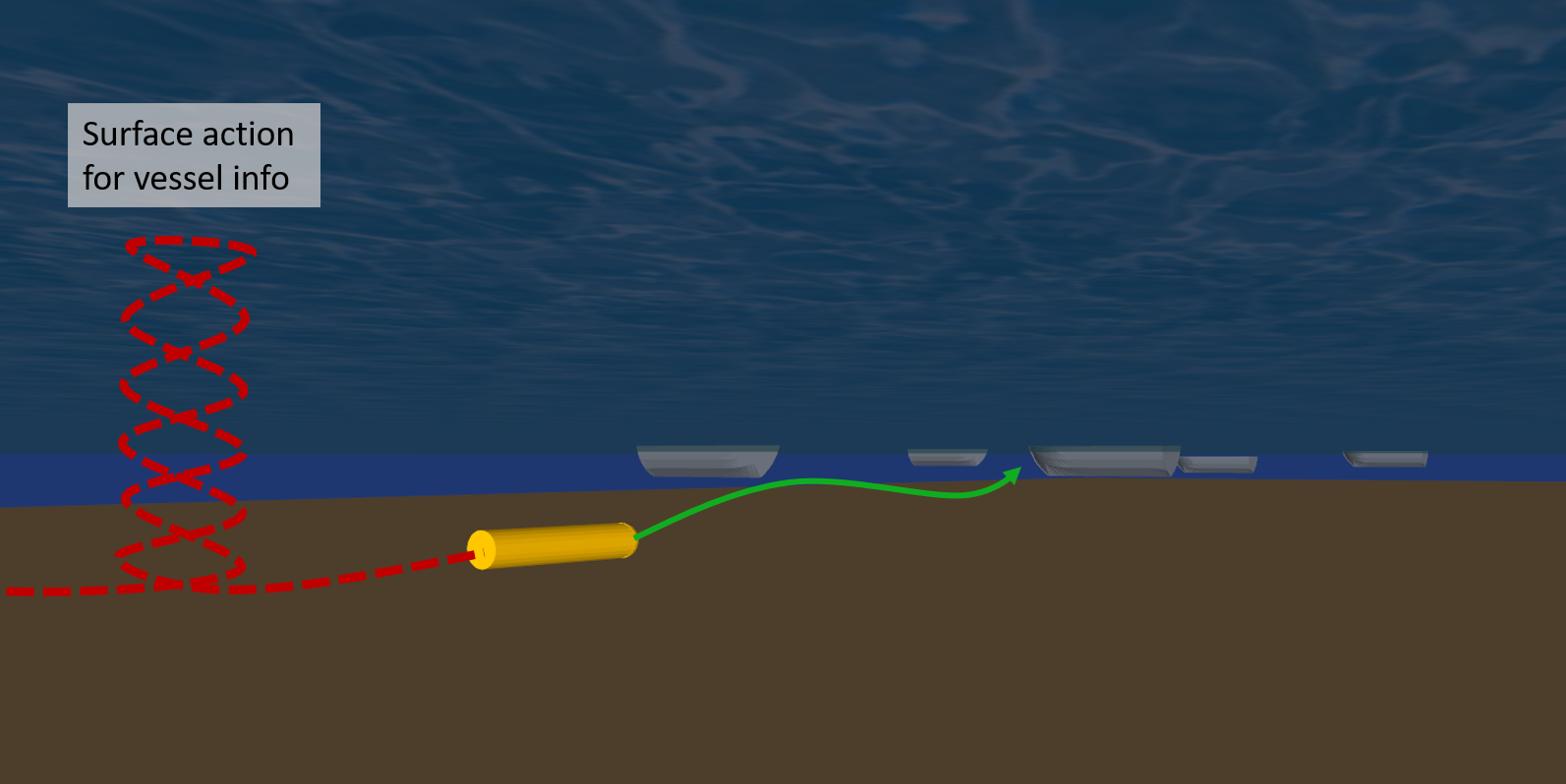}
        \vspace{-15pt}
    \caption{Pictorial representation of a UUV tasked to navigate through a shipping channel under intermittent communication.}
    \label{fig:intro}
    \vspace{-20pt}
\end{figure}

Safe motion and task planning for multi-robot systems in known and static environments has been studied extensively under the assumption that all robots are connected and synchronously exchanging information about their local states and actions \cite{van2008}. However, when robots are deployed in communication- and sensor-limited dynamic environments, data about the current state of the environment may not always be available. For example, consider unmanned underwater vehicles (UUVs), which have been growing in popularity in recent years \cite{Fletcher2000}. UUVs must deal with vessels that typically are not aware of their presence in the water. In fact, UUVs obtain information about the states of the other vessels only when they surface; see, e.g., Fig.~\ref{fig:intro}. Hence, when they are underwater they can only rely on old, deprecated data to plan their motion to avoid colliding with these vessels. This sensory limitation may threaten the safety of such systems. For example, on February 8, 2021, a Japanese submarine collided with a commercial ship off the country's Pacific coast.
The submarine surfaced right under the ship, unaware of its presence, and suffered damage to its diving and communication capabilities. More generally, this type of problem also appears when robots are operating in communication/sensor limited environments, like unmanned aerial or ground vehicles operating in cluttered and adversarial environments (e.g., during military operations).

In this paper, we address the problem of controlling a mobile robot to safely navigate communication- and sensor-limited dynamic environments. The environment considered is (i) \textit{dynamic} in the sense that it is occupied with non-cooperative mobile obstacles/vehicles that move according to known but noisy dynamics;  and (ii) \textit{communication/sensor limited} in the sense that the robot cannot sense or communicate with other mobile obstacles/vehicles. We assume that the robot receives intermittent information about the obstacle locations and knows obstacle dynamics and control policies which are subject to unknown process noise.
%
%
%
The major challenge of safe navigation in such environments lies in the intermittent nature of the data related to other vehicles, which requires the robot to plan over stale information. To safely plan with stale information, the robot must reason about the future dynamic obstacle configurations while taking their motion uncertainties into consideration until fresh data are available.
%

%
%
To address this challenge, 
%
we design a novel reachability-based potential field method to avoid collisions with future obstacle configurations. Specifically, reachability analysis is employed to compute the future states of obstacles based on the most recent information the robot received while potential fields are used to avoid these reachable sets. Since reachability analysis is too computationally expensive to perform at runtime, we use this reachability-based potential field method to train
a neural network (NN) control framework at design time. At runtime, given intermittent information, this NN component generates control actions to safely navigate the dynamic environment.

\textbf{Related work:} 
Motion planning in dynamic environments has been addressed through methods such as covariant hamiltonian optimization for motion planning 
\cite{Zucker2013,Byravan2014,Men2020}, real-time adaptive motion planning 
\cite{Vannoy2008,Mcleod2016,guzzi2013human,aoude2013probabilistically,renganathan2020towards} and reinforcement learning \cite{Liu2020}. However, these methods require real-time sensor information about the dynamic obstacles. Dimension reduction methods have also been proposed \cite{Vemula2016}, but they do not handle noise in the obstacle dynamics.

Relevant approaches on motion planning under intermittent information have been proposed recently. For instance, \cite{Khodayi2019,Kantaros2019,aragues2021intermittent} propose distributed intermittent connectivity controllers for multi-robot systems that are tasked with navigation and estimation tasks in communication-denied environments. Conditions on the time interval between intermittent communication events to ensure stability and consensus in multi-agent systems are also developed in \cite{Xu2019}. Similar to this work, mission planning problems under uncontrolled intermittent communication have recently been proposed as well in which localization \cite{Bopardikar2016,Penin2019,Yel2019} and object tracking \cite{Koohifar2018} rely on intermittent information. The above works consider cooperative multi-agent systems whereas our work considers robots navigating in environments occupied with uncertain, dynamic, and uncooperative obstacles. Additionally, reachability analysis has been employed for control design \cite{Seo2019,Ding2011,Malone2017,Pendleton2017,Desai2020, Akametalu2015, Chiang2017}, monitoring \cite{Yel2019,Franco2020}, and providing run-time safety guarantees \cite{Althoff2014,Herbert2019}. The most relevant works to ours are: \cite{Malone2017}, where reachability analysis and potential fields are mixed for planning, and \cite{Chiang2017}, which combines stochastic reachability analysis and RRT for path planning. Differently from our work, \cite{Malone2017} assumes periodic observations which doesn't address intermittent information challenges and \cite{Chiang2017} does not account for the computational complexity of stochastic reachability analysis.



\textbf{Contribution:} The contribution of this paper is four-fold: 1) we present a novel algorithm that combines reachability analysis (RA) with potential fields for safe navigation in uncertain and dynamic environments in the presence of intermittent information; 2) we propose an NN-based compositional method that leverages the time elapsed since the last intermittent data to perform fast planning bypassing the use of expensive RA at runtime; 3) we implement and validate our approach with realistic simulations and experiments on UUVs and ground vehicles in cluttered dynamic environments; 4) we use comparative experiments against planning methods in dynamic environments that ignore motion uncertainties to demonstrate the safety-related benefits of our approach.

\section{Problem Description}
\label{sec:formulation}

Consider a robot governed by the following dynamics:
\begin{equation}\label{eq:rdynamics}
\dot{\bm{x}}=f(\bm{x}(t),\bm{u}(t)),
\end{equation}
where $\bm{x}(t)$ and $\bm{u}(t)$ denote the state (e.g., position and heading) of the robot and the control input at time $t$, respectively. The robot is assumed to operate in an environment with an a priori unknown number $n\geq0, n\in\mathbb{N}$ of dynamic, non-cooperative obstacles governed by the following dynamics:
\begin{equation}\label{eq:odynamics}
\dot{\bm{o}}_i=g(\bm{o}_i(t),\bm{d}_i(t),\bm{w}_i(t)), \;\; i=1,\ldots,n
\end{equation}
where $\bm{o}_i(t) \in \mathcal{O}$, $\bm{d}_i(t)\in \mathcal{D}$ and $\bm{w}_i(t) \in \mathcal{W}$ denote the state, control input and process noise of obstacle $i$ at time $t$, respectively. Hereafter, we assume that all obstacles follow the same control policy, which is known to the robot. 
In addition, the system noise is considered unknown but bounded by an a priori known bound $\lVert \bm{w}_i(t) \rVert \leq \delta_w$.

The key assumption of this work is that the robot gets intermittent information about the obstacle state $\bm{o}_i(t_s)$,
which could be planned (e.g., when the UUV surfaces) or at unknown time instants $t_s$ (e.g., a UGV entering a communication denied environment).

The problem we address in this paper is as follows:

\begin{problem}[Safe Planning with Intermittent Information]\label{problem}
Given a robot with dynamics in \eqref{eq:rdynamics}, design a planner to guide it to a goal region ${\bm{x}}_g$ while always avoiding dynamic vehicles modeled as in \eqref{eq:odynamics}. This is specified in the following safety condition: 
 \begin{align}\label{eq:safety}
        \left \lVert \bm{x}(t) - \bm{o}_{i}(t)  \right \rVert \geq \delta, \; \; \forall t\geq0, \; \forall i \in \{ 1,\hdots,n\}
\end{align}
with safety threshold $\delta$ under intermittent obstacle-related information $\mathcal{I}(t_s)$ provided at unknown time instants $t_s$.

\end{problem}

\section{Planning Framework}
\label{sec:approach}

\begin{figure}[t]
    \centering
    \includegraphics[width=0.49\textwidth]{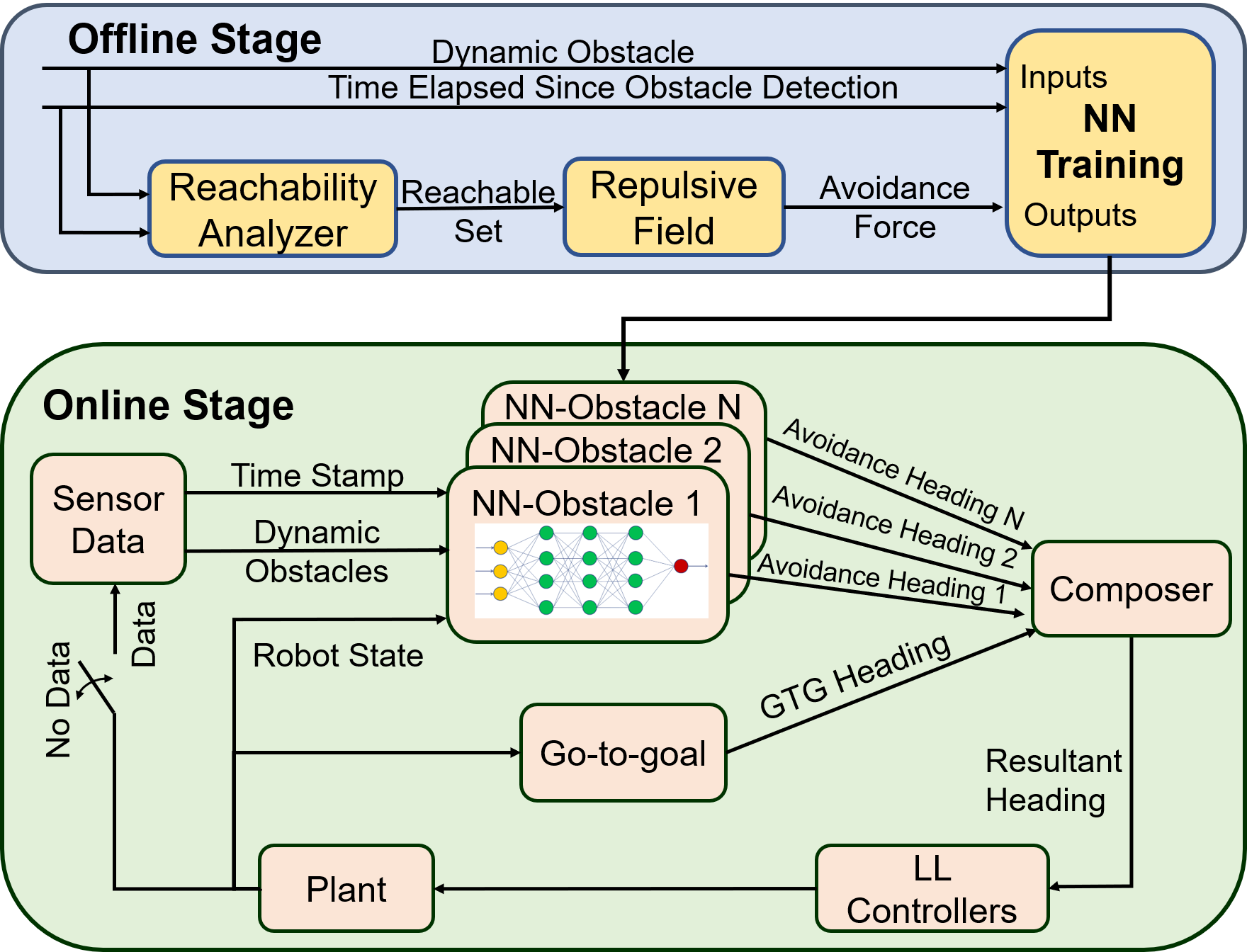}
    \caption{Architecture of our proposed approach.}
    \label{fig:highLevelApproach}
    \vspace{-10pt}
\end{figure}

To solve Problem 1, we propose to leverage (i) reachability analysis, which allows us to predict future states of the obstacles in the presence of uncertainty during the periods between intermittent information and (ii) potential fields to navigate the robot in the environment while avoiding the reachable sets associated with the obstacles. Nevertheless, a key challenge in employing reachability analysis (RA) is that
it is computationally expensive, resulting in limited applicability in online operations. To mitigate this challenge, we consider a learning-based approach leaving RA offline and enabling fast online planning and control. 
The proposed learning-based framework is summarized in Fig.~\ref{fig:highLevelApproach} and consists of 1) an offline phase where a neural network (NN) controller is trained under RA and potential fields and 2) an online phase in which the NN controller, given the most recently received data, generates safe control actions to quickly navigate the dynamic environment. In what follows, we describe in detail the components of the proposed method depicted in Fig.~\ref{fig:highLevelApproach}.
%
\subsection{Reachability Analysis for Dynamic Obstacles} \label{sec:reachability}

In this section, we discuss how reachability analysis methods can be used to predict future states of the dynamic obstacles.
In particular, given a dynamic obstacle $i$, as defined in \eqref{eq:odynamics}, along with a set of initial conditions $\mathcal{O}_{t_0}^i$ at time $t_0$, our goal is to compute the set of possible states that obstacle $i$ can be in from a time instant $t_1 \geq t_0$ until a time instant $t_2 \geq t_1$.
Hereafter, we call such sets reachable sets, denote them by $\mathcal{O}_{[t_0,t_1,t_2]}^i$, and define them as follows:
\begin{align} \label{eq:RA}
    \text{$\mathcal{O}_{[t_0,t_1,t_2]}^i$} = \{ \bm{o} \in \mathcal{O} \mid &~\exists \bm{o}_{0} \in \mathcal{O}_{t_0}^i, \text{$t' \in [t_1,t_2]$}, \text{$\bm{d}_{[t_0,t']}$} \in \mathcal{D}, \\
    \text{$\bm{w}_{[t_0,t']}$} \in \mathcal{W},
    &~\bm{o}=G_{t_0,\text{$t'$}}(\bm{o}_0,\bm{d}_{[t_0,t']},\bm{w}_{[t_0,t']})\}, \nonumber
\end{align}
where $G_{t_0,\text{$t'$}}(\mathcal{O},\mathcal{D},\mathcal{W})$ is the system's transition function (i.e., the state the system will be in at time $t'$ when it is in state $\bm{o}_{0}$ at time $t_0$ and is applied input signal $\bm{d}_{[t_0,t']}$ and process noise signal $\bm{w}_{[t_0,t']}$). There exist various methods to compute the reachable set \eqref{eq:RA}, such as Taylor models \cite{Chen2013}, Hamilton-Jacobi theory \cite{Bansal2017}, $\delta$-reachability \cite{Soonho2015}, and ellipsoids \cite{Kurzhanskiy2006}. Our framework is agnostic to the method used, provided that we can extract the reachable sets over time.

\subsection{Potential Fields for Navigating Dynamic Environments}\label{sec:potentialfields}


In what follows, we design potential field controllers to navigate the robot through dynamic environments. The key idea is to design a controller capable of avoiding all reachable obstacle states from current time $t$ until the next batch of information arrives, denoted as $\mathcal{O}_{[t_s,t,t_{s+1}]}$, and therefore, the dynamic obstacles within the time interval $[t,t_{s+1}]$; recall that $t_s$ and $t_{s+1}$ stand for consecutive time instants where the robot receives information about the obstacles.

To design a potential field controller for dynamic environments, we first define the following potential field:
\begin{align}
    U(\bm{x}_p) = U_{\text{att}}(\bm{x}_{p})+\sum_{i=1}^{n} U_{i,\text{rep}}(\bm{x}_{p}), \label{eqn:composedField}
\end{align}%
where $U_{\text{att}}(\bm{x}_{p})$ denotes an attractive field driving the robot towards the goal location $\bm{x}_{g}$ and $U_{i,\text{rep}}(\bm{x}_{p})$ is a repulsive field pushing the robot away from dynamic obstacle $i$. The attractive field is designed to take small values near the robot's goal and large values far from the goal while the repulsive field is constructed to take large values near the obstacles and small values far from them. The control policy is to simply perform gradient descent down this field until the goal is reached.

Letting $\bm{x}_{p}$ denote the current robot position, the attractive field is defined as:
\begin{align}
    U_{\text{att}}(\bm{x}_{p}) = \frac{1}{2}K_p \left (\lvert \lvert \bm{x}_p-\bm{x}_{g}\rvert \rvert^2 \right).
    \label{eqn:pfeqnAtt}
\end{align}

\begin{figure}[t]
\centering
 \subfigure[]{\includegraphics[width=0.23\textwidth]{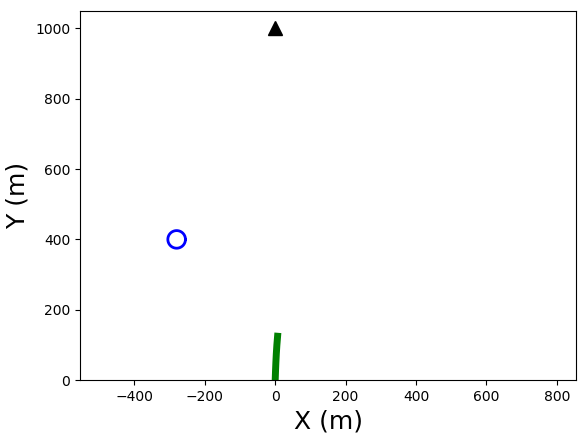}\label{fig:mean and std of net14}}\hspace{1em}%
 \subfigure[]{\includegraphics[width=0.23\textwidth]{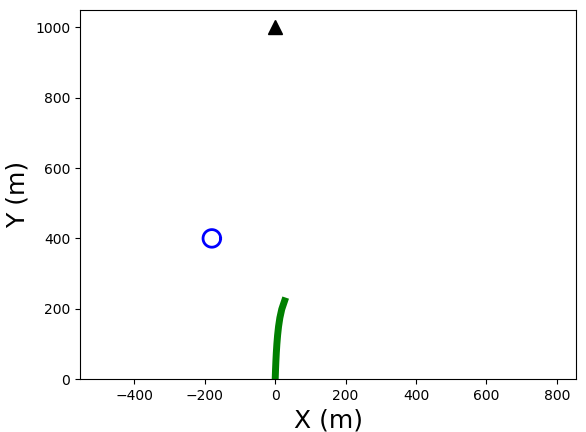}\label{fig:mean and std of net24}}
  \subfigure[]{\includegraphics[width=0.23\textwidth]{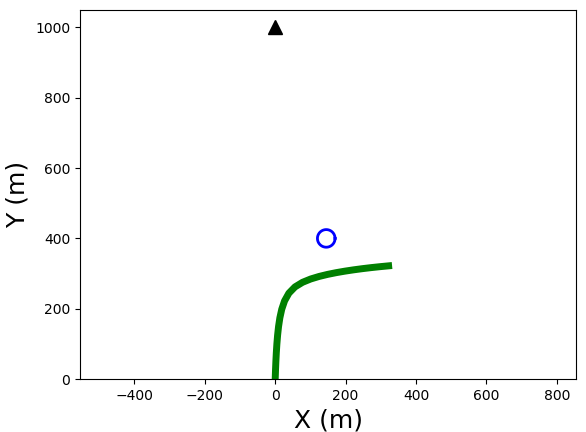}\label{fig:mean and std of net34}}\hspace{1em}%
 \subfigure[]{\includegraphics[width=0.23\textwidth]{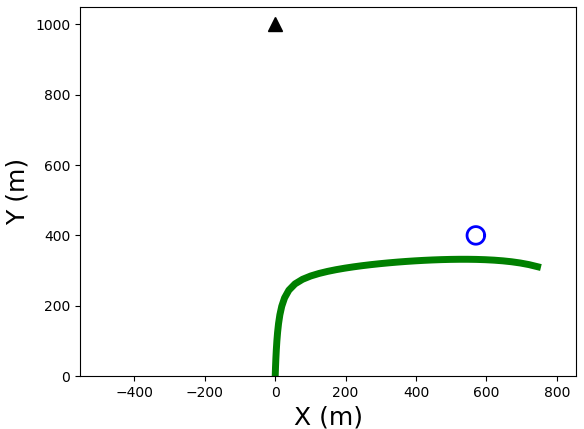}\label{fig:mean and std of net44}}\label{fig:mean and std of net44}
\caption{Example of a dynamic minima in potential field planning. The robot's path is in green, the goal point is in black, and the obstacle is the blue circle.}
\label{fig:dynamicMinima}
\vspace{-10pt}
\end{figure}

\begin{figure}[ht!]
    \centering
    \includegraphics[width=\linewidth]{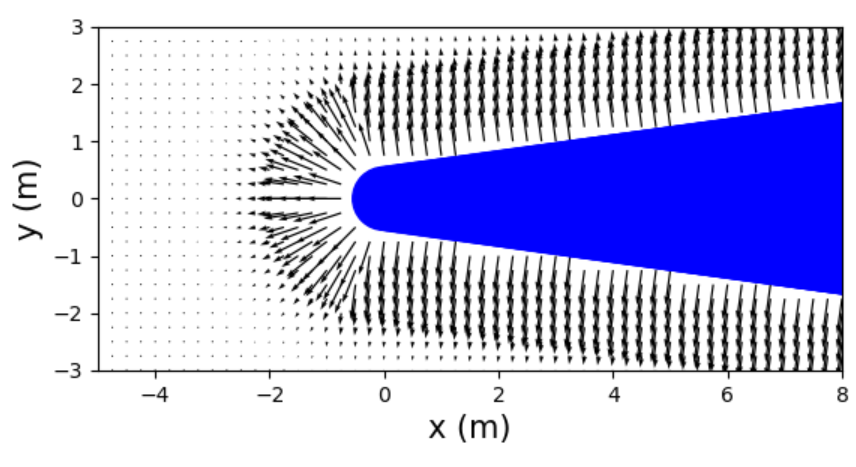}
        \vspace{-20pt}
    \caption{Repulsive forces from the forward reachable set (blue) of an obstacle.} 
    \label{fig:repFieldOfObst}
    \vspace{-10pt}
\end{figure}

Since the control action is to follow the negative gradient of the composed field, potential field controllers can make the system get stuck in \textit{dynamic minima}. These occur when a dynamic obstacle repels the robot in the same direction that the obstacle is moving; see, e.g. Fig.~\ref{fig:dynamicMinima}. To minimize the effects of dynamic minima, we leverage the temporal properties of reachability analysis. Every time obstacle information is provided to the robot, we construct the obstacles' forward reachable set defined in \eqref{eq:RA}. Exploiting the conic structure of the reachable sets, we design repulsive potential fields that tend to push the robot behind the obstacle, thus mitigating the dynamic minima behavior seen in Fig.~\ref{fig:dynamicMinima}. For instance,  Fig.~\ref{fig:repFieldOfObst} shows an example of the repulsive forces from the reachable set of an obstacle starting at position $[0,0]$ and moving in the positive $x$ direction. 

We formally describe this process for obstacle $\bm{o}_i$ as follows: Let $t$ be the current global time, $\bm{x}_p$ be the location of the robot at the current time, $t_0 \leq t$ be the time of the most recently received obstacle state, denoted $\bm{o}_{t_0}^i$. Also, let $\mathcal{O}^{i}_{[t_0,t,t_0+T]}$ be the set of reachable states of $\bm{o}_i$ from time $t$ to $t_0+T$ (see \eqref{eq:RA}), given that $\bm{o}_i$ was at state $\bm{o}_{t_0}^i$ at time $t_0\leq t$.

Finally, define $\bar{\bm{o}}^{i}_{t}(\bm{x}_p)$ to be the closest point in $\mathcal{O}^{i}_{[t_0,t,t_0+T]}$ to the robot's position $\bm{x}_p$:
\begin{align}
    \bar{\bm{o}}^{i}_{t}(\bm{x}_{p}) \coloneqq \text{argmin}_{\bm{o} \in \text{$\mathcal{O}^{i}_{[t_0,t,t_0+T]}$}} \left \lVert \bm{x}_{p} - \bm{o} \right \rVert.
\end{align}

The repulsive field that we use to compute the avoidance command for obstacle $\bm{o}_{i}$ at robot location $\bm{x}_p$ and time $t$ is:

\begin{align}
    U_{t,\text{rep}}^{i}(\bm{x}_{p}) = \frac{1}{2} K_{r} \left(\frac{1}{\left \lVert \bm{x}_{p} - \bar{\bm{o}}^{i}_{t}(\bm{x}_{p}) \right \rVert -\delta} \right)^2, \label{eqn:augmentedRepFieldSingleObst}
\end{align}
where $\delta$ is given from \eqref{eq:safety}.

\subsection{Compositional Controller}
\label{sec:compositionAndGuarantees}
The final step for our control scheme is composing the attractive and repulsive fields. 
%
%
%
By combining the attractive \eqref{eqn:pfeqnAtt} and repulsive \eqref{eqn:augmentedRepFieldSingleObst} field equations, based on the composition defined in \eqref{eqn:composedField}, the composed field becomes:
\begin{align}
     U_{t}(\bm{x}_p) &=  \left ( \frac{1}{2}K_p \left (\lvert \lvert \bm{x}_{p}-\bm{x}_{g}\rvert \rvert^2 \right ) \right ) + \nonumber \\
    & \; \; \sum_{i=1}^n \frac{1}{2} K_{r} \left(\frac{1}{\left \lVert \bm{x}_{p} - \bar{\bm{o}}^{i}_{t}(\bm{x}_{p}) \right \rVert -\delta}\right)^2. \label{eqn:fullfield}
\end{align}

Our controller uses the same principle of following the negative gradient down the composed potential field as the standard potential field setup. Using the linearity of the gradient, the control output becomes:\footnote{This is standard in potential field planning but it is critical for the compositionality of our approach, so we write out these steps.}
\begin{align}
    \bm{u} = &- \nabla \left(\frac{1}{2}K_p \left (\lvert \lvert \bm{x}_{p}-\bm{x}_{g}\rvert \rvert^2 \right) \right) - \nonumber \\
    & \; \; \sum_{i=1}^n \nabla \frac{1}{2} K_{r} \left(\frac{1}{\left \lVert \bm{x}_{p} - \bar{\bm{o}}^{i}_{t}(\bm{x}_{p}) \right \rVert -\delta}\right)^2. \label{eqn:fullgrad}
\end{align}

Next, we discuss intuitively under what conditions the controller in \eqref{eqn:fullgrad} satisfies the safety constraint \eqref{eq:safety}.

\vspace{5pt}
{\em{Safety Discussion:}} First, note that if the reachable set of obstacle $i$ from times $t$ to $t_0+T$, $\mathcal{O}^{i}_{[t_0,t,t_0+T]}$, is convex, the repulsive field vector produced by that obstacle, $\nabla \frac{1}{2} K_{r} \left(\frac{1}{\left \lVert \bm{x}_{p} - \bar{\bm{o}}^{i}_{t}(\bm{x}_{p}) \right \rVert -\delta}\right)^2$, points away from every point of the obstacle's forward reachable set.\footnote{This can be shown by contradiction. Assume that there is a point of the reachable set which the repulsive field points towards. Then it is not possible for the reachable set to be convex.}

For a collision to occur, the repulsive field $U_{t}(x_p)$ must take an infinite value, since the distance from the robot to the obstacle appears in the denominator of each obstacle's repulsive field. So to prove safety it is sufficient to show that $U_{t}(x_p)$ is always upper bounded.

At time $t$, as the robot approaches the reachable set of obstacle $i$, denoted by $\mathcal{O}^{i}_{[t_0,t,t_0+T]}$, the repulsive force generated by the reachable set will become so large as to render the forces from every other obstacle reachable set (assuming the distance between the reachable sets is large enough) and the goal negligible. Thus, as the robot approaches the reachable set of obstacle $i$, it holds that:

\begin{align}
        U_{t}(x_p) &\approx \frac{1}{2} K_{r} \left(\frac{1}{\left \lVert \bm{x}_{p} - \bar{\bm{o}}^{i}_{t}(\bm{x}_{p}) \right \rVert -\delta} \right)^2 \\
        u &\approx \nabla \frac{1}{2} K_{r} \left(\frac{1}{\left \lVert \bm{x}_{p} - \bar{\bm{o}}^{i}_{t}(\bm{x}_{p}) \right \rVert -\delta}\right)^2
\end{align}

Assuming simple robot dynamics of the form $\dot{{\bm{x}}}(t)={\bm{u}}$, as e.g. in \cite{Vasilopoulos2018}, this control command pushes the robot farther away from every point of $\mathcal{O}^{i}_{[t_0,t,t_0+T]}$. In addition the reachable sets of the obstacles do not grow, since $\mathcal{O}^{i}_{[t_0,t,t_0+T]} \supseteq \mathcal{O}^{i}_{[t_0,t',t_0+T]}, ~\forall t' \geq t$. Thus $U_{t}(x_p)$ is decreasing and is hence upper bounded.



By construction of the reachable sets, the above argument assures safety up to a time threshold $T$, which depends on how far the forward reachable sets get extended. If $T=\infty$ then the safety guarantee holds for all time and as $T$ decreases safety gets sacrificed, but the controller produces less conservative paths. In practice, this threshold should be picked to ensure safety until the next batch of intermittent information arrives. This threshold also gives a way to balance having conservative or aggressive paths.

\subsection{Compositional Neural Network Controller} \label{sec:neural_net}

\begin{figure*}[h!]
    \centering
     \subfigure[]{\includegraphics[width=0.23\textwidth]{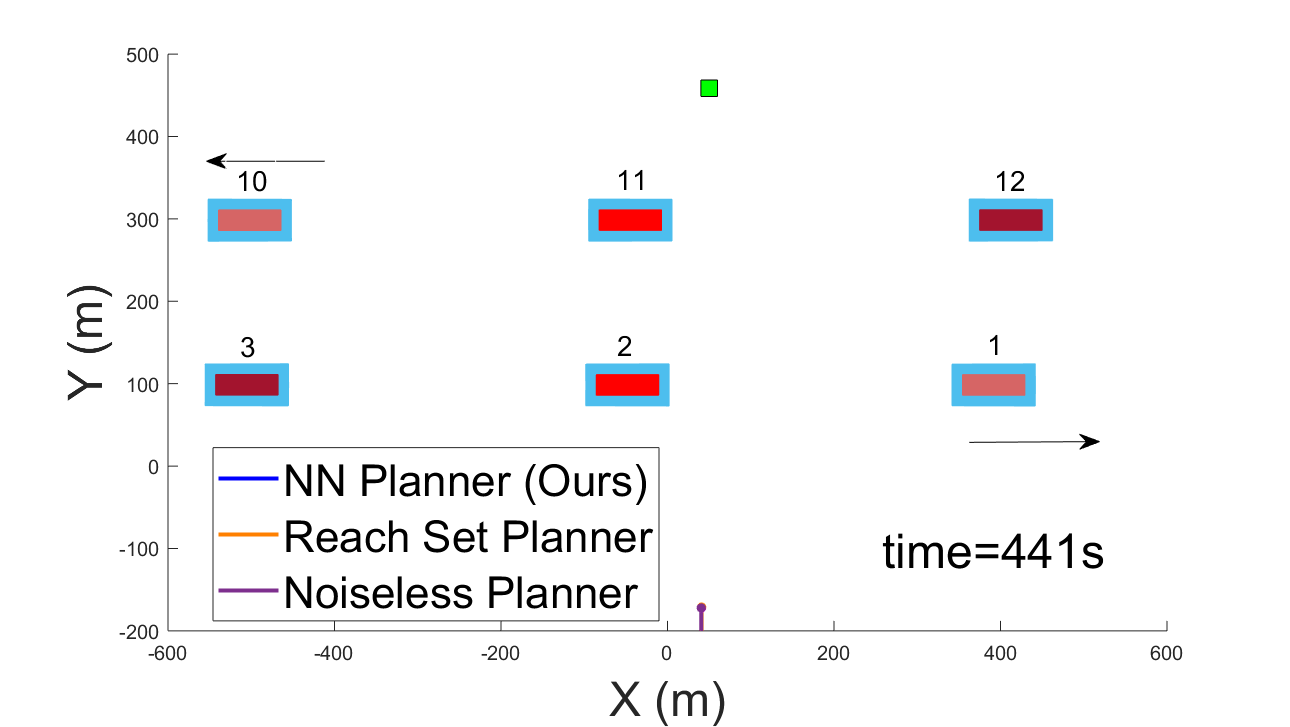}\label{fig:uuvpath1}}\hspace{1em}%
     \subfigure[]{\includegraphics[width=0.23\textwidth]{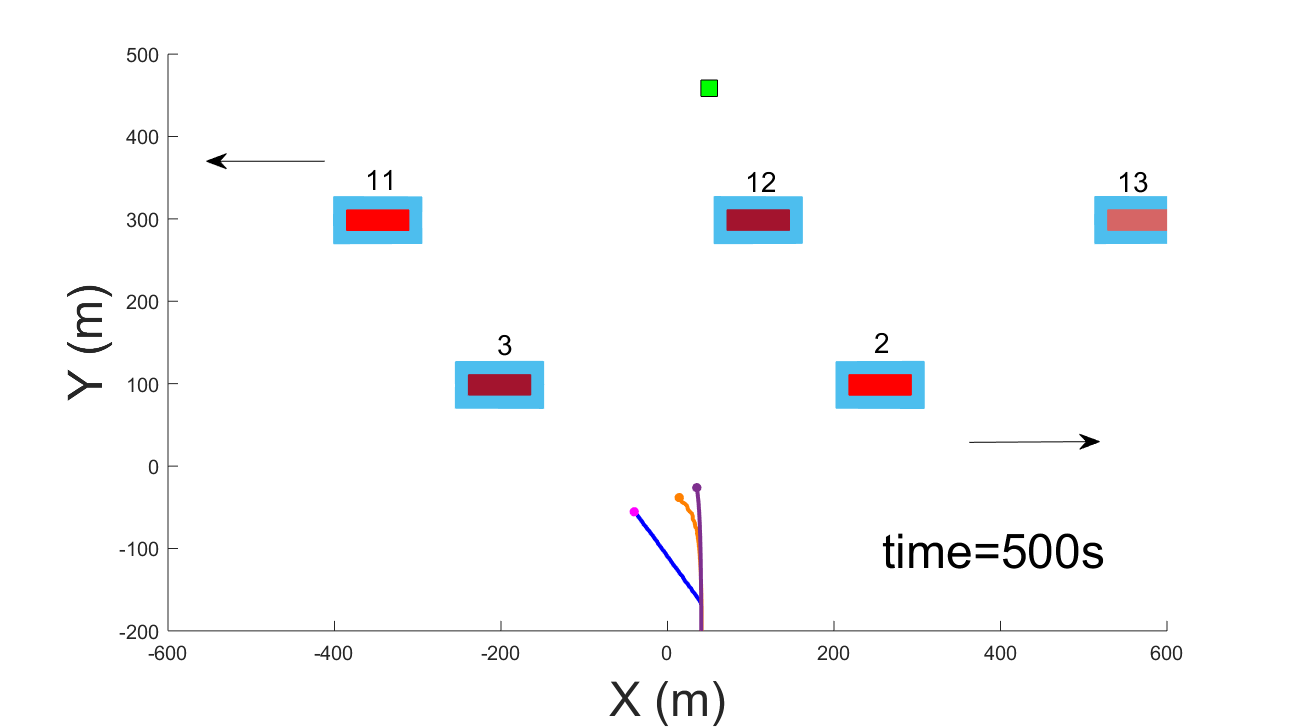}\label{fig:uuvpath2}}
     \subfigure[]{\includegraphics[width=0.23\textwidth]{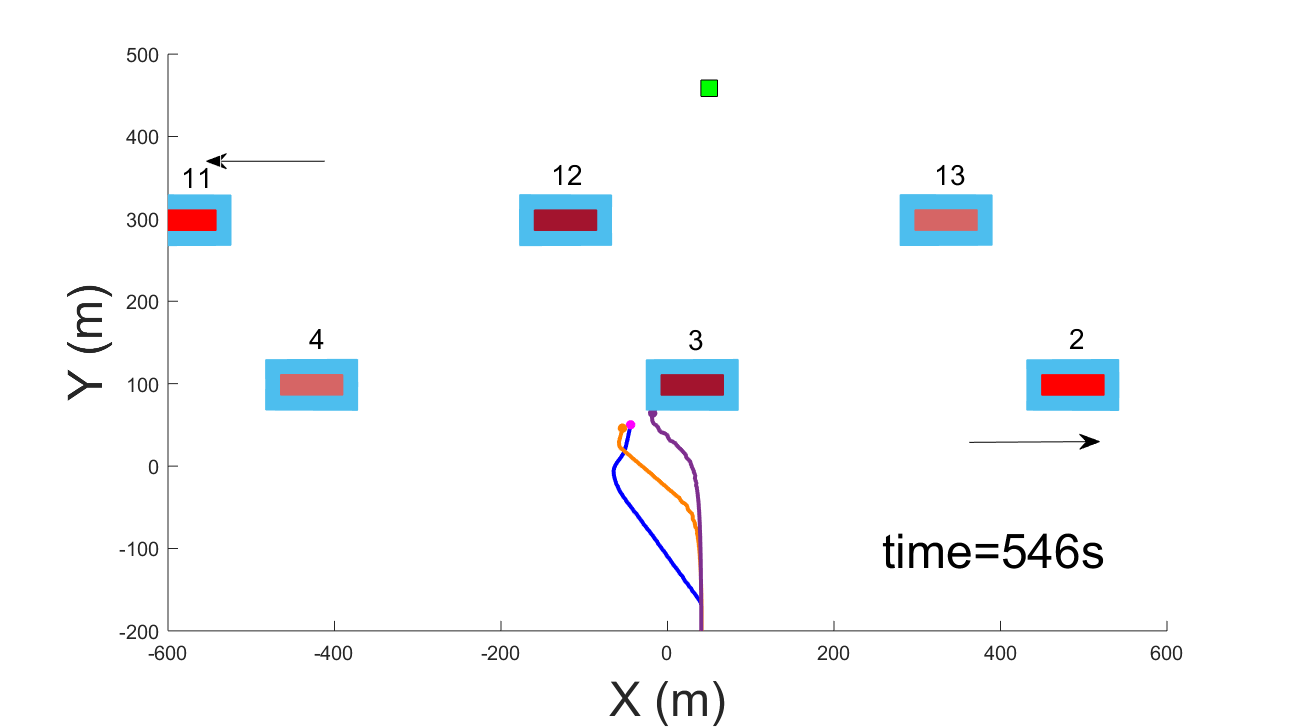}\label{fig:uuvpath3}}\hspace{1em}%
     \subfigure[]{\includegraphics[width=0.23\textwidth]{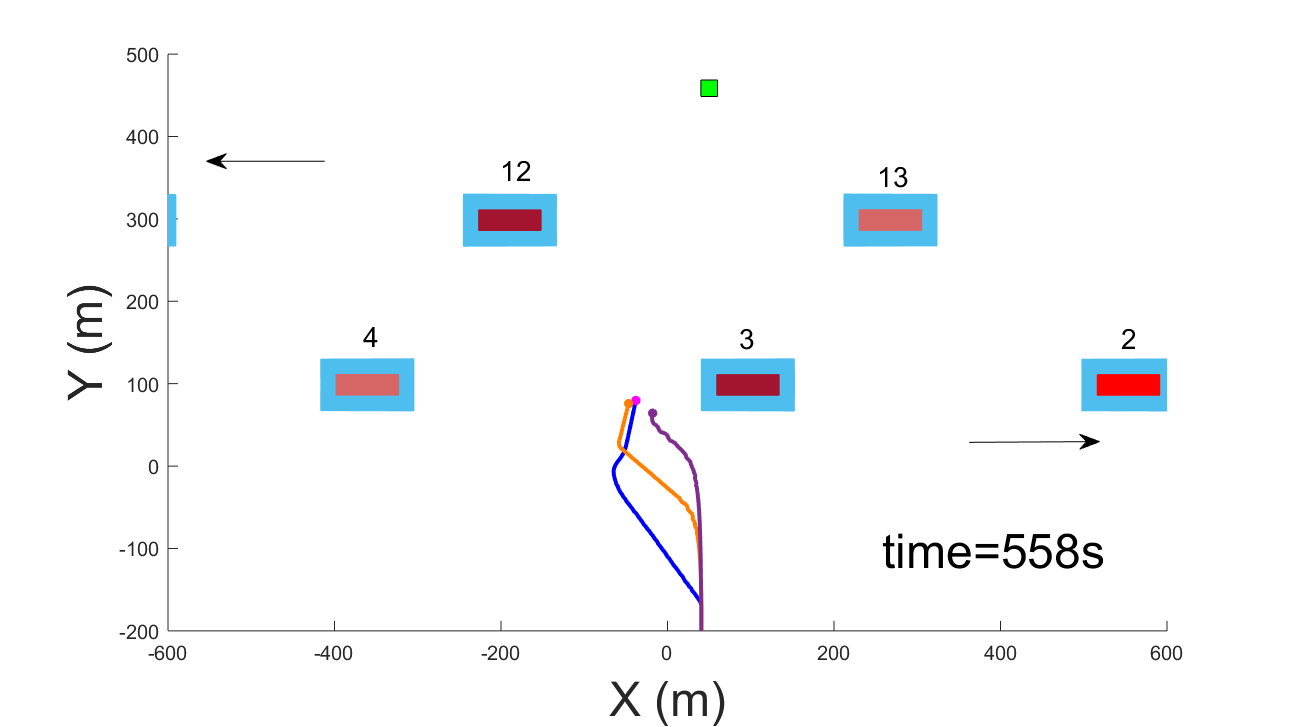}\label{fig:uuvpath4}}
     \subfigure[]{\includegraphics[width=0.23\textwidth]{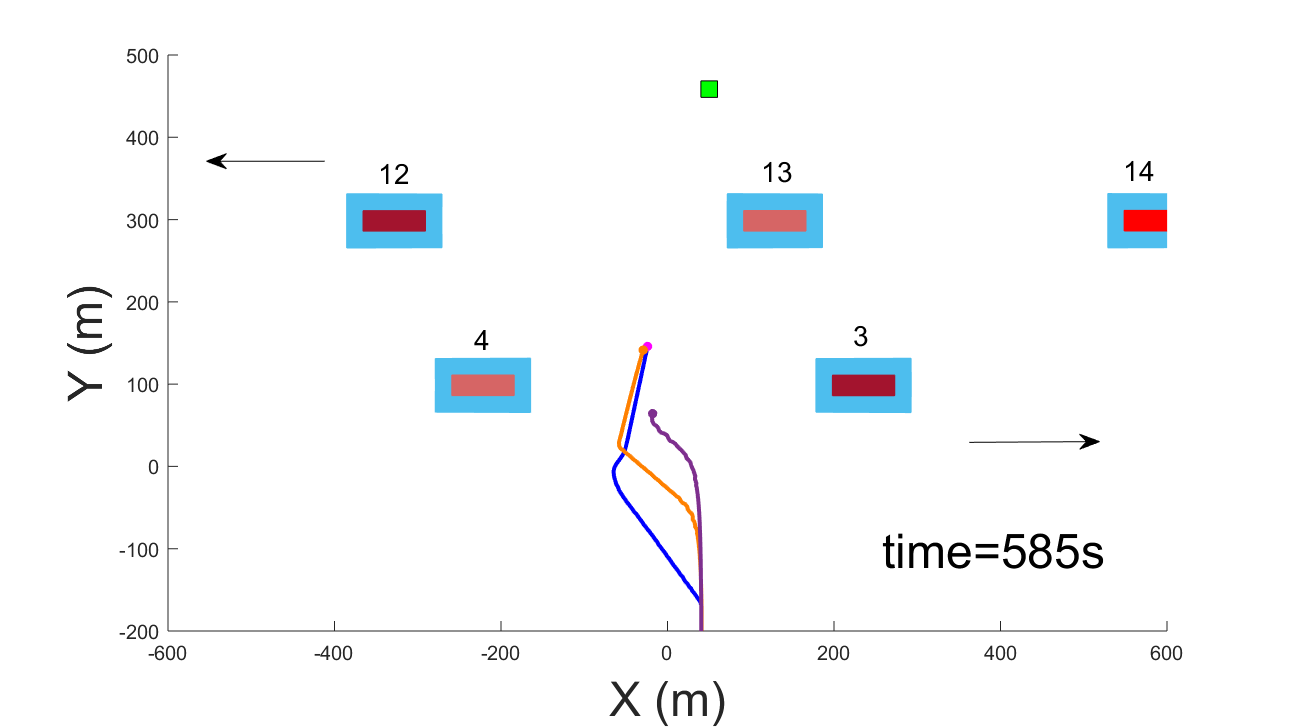}\label{fig:uuvpath5}}\hspace{1em}%
     \subfigure[]{\includegraphics[width=0.23\textwidth]{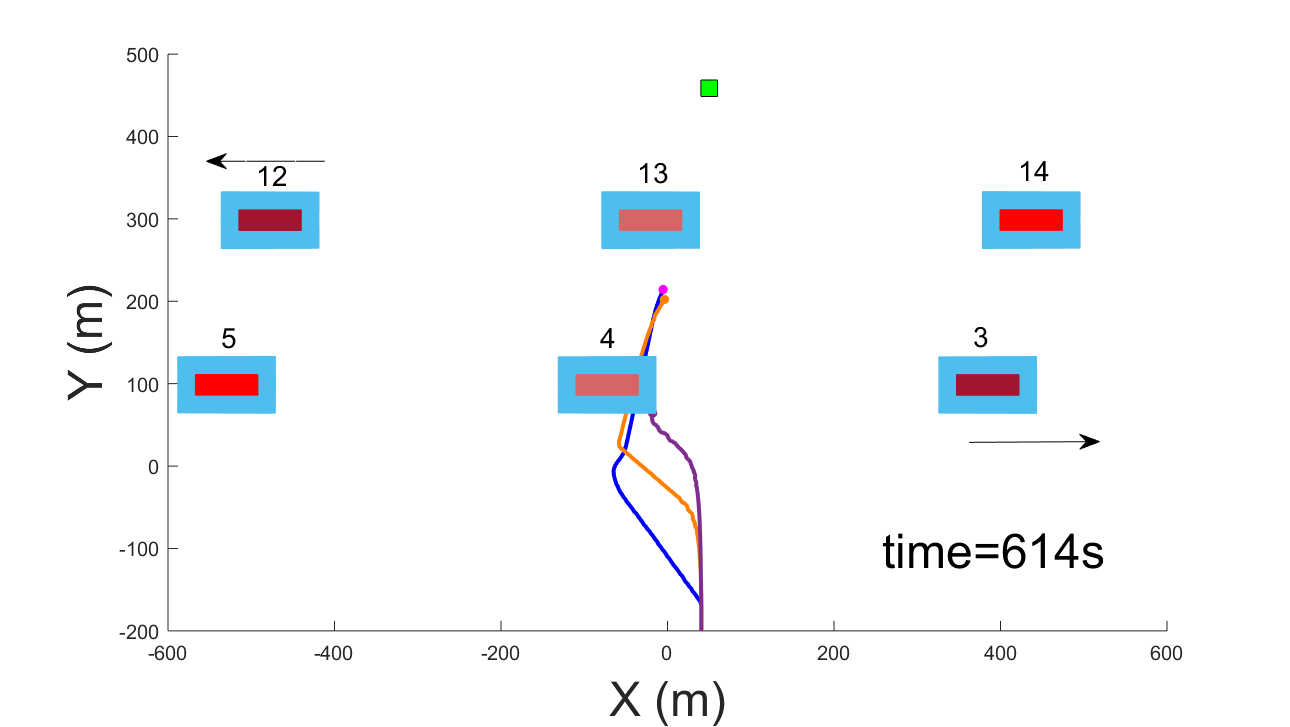}\label{fig:uuvpath6}}
     \subfigure[]{\includegraphics[width=0.23\textwidth]{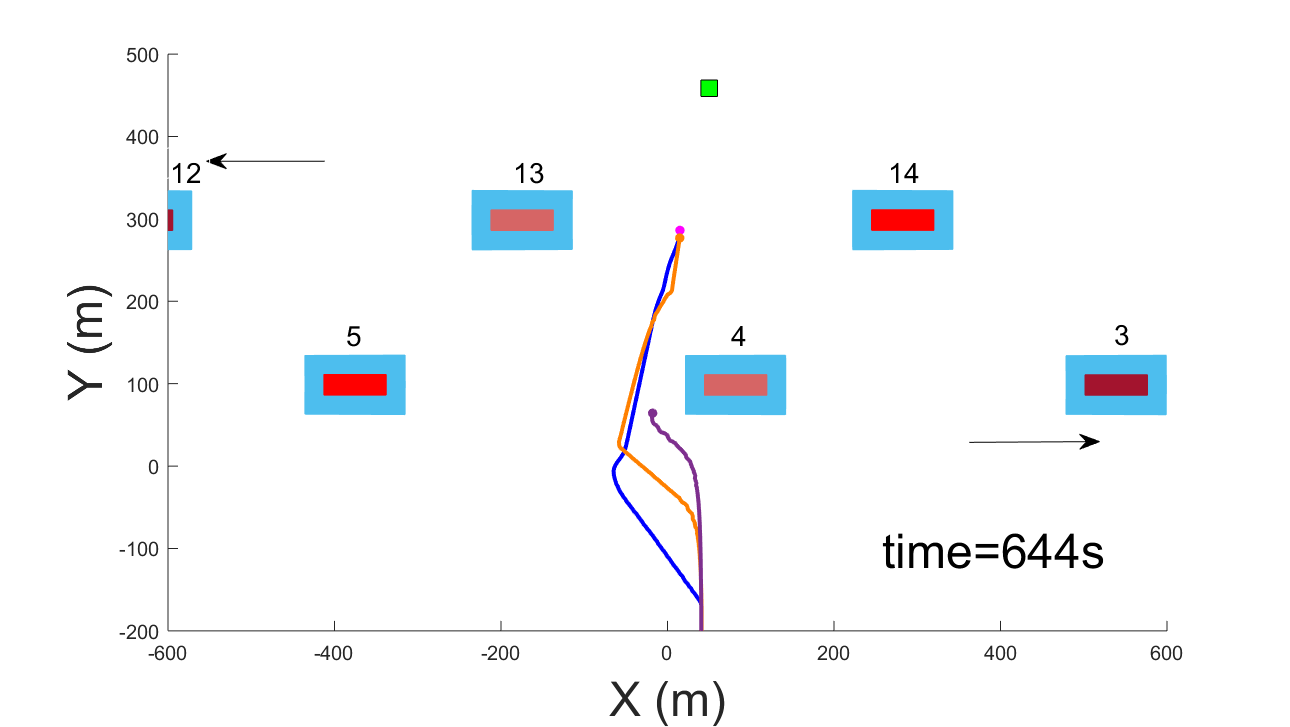}\label{fig:uuvpath7}}\hspace{1em}%
     \subfigure[]{\includegraphics[width=0.23\textwidth]{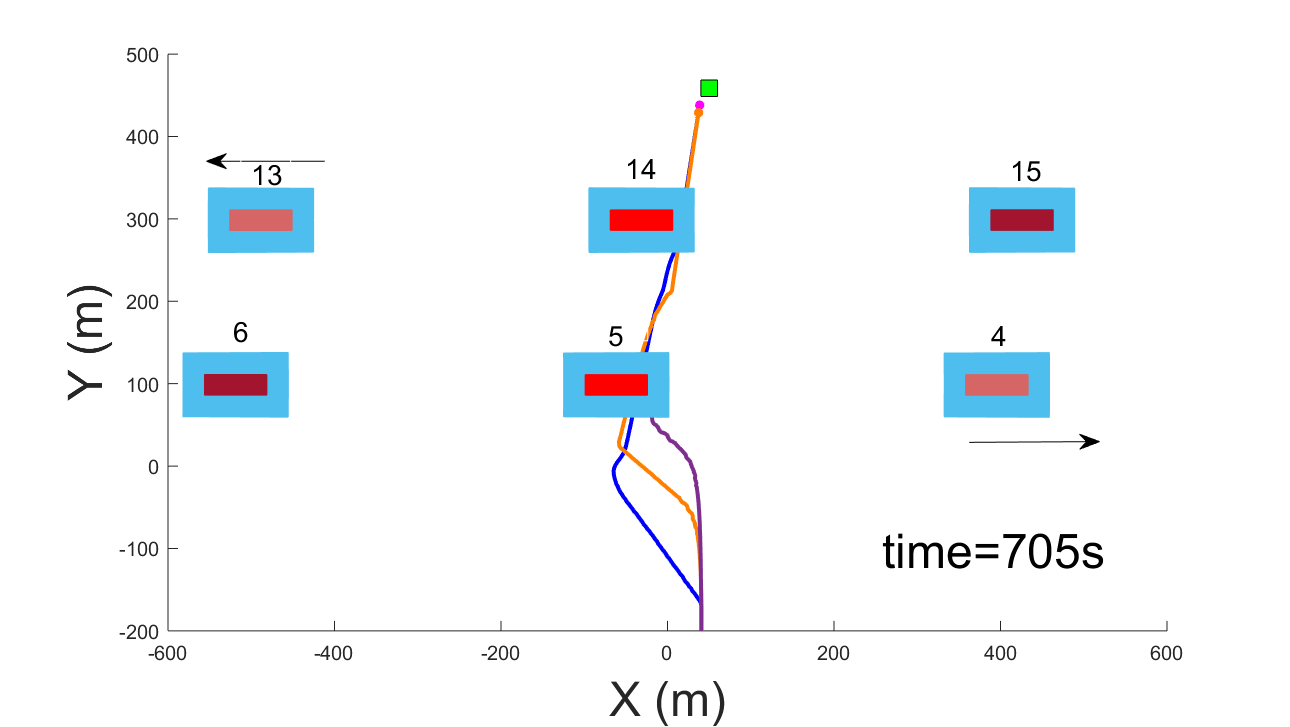}\label{fig:uuvpath8}}
    \vspace*{-3mm}
    \caption{UUV crossing a shipping channel with ship velocities $5.14$m/s. Each ship is represented as a red box, each ship reachable set is represented as a light blue square, and the goal point is the green box. }
    \label{fig:UUVpath}
    \vspace{-10pt}
\end{figure*}

\begin{figure}[!h]
    \centering
    \includegraphics[width=\linewidth]{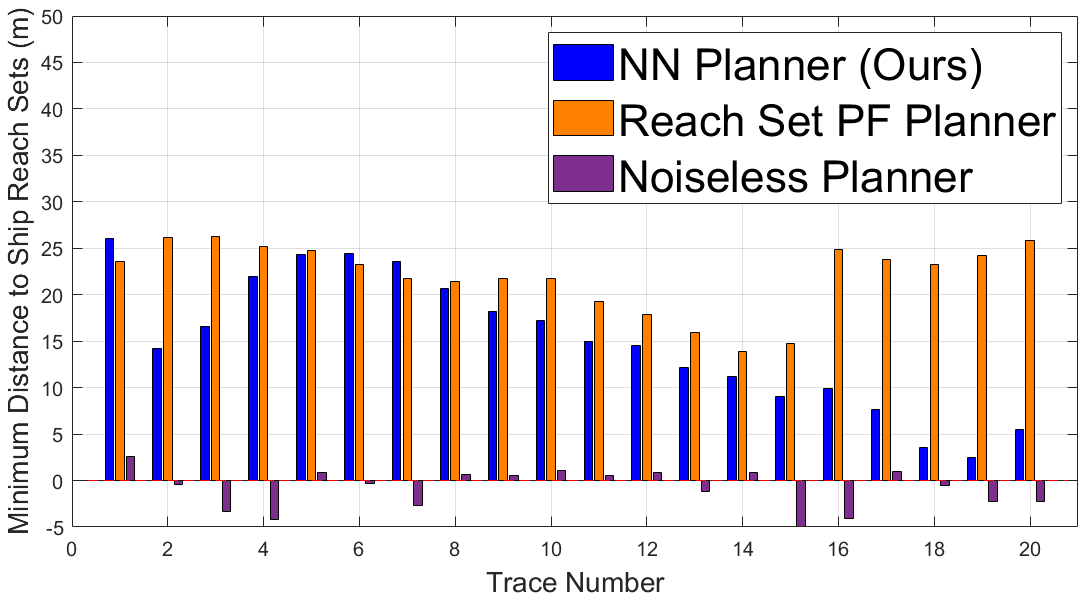}
    \caption{Minimum distance between the UUV and the ship reachable sets across all $20$ trials, shifted so that the y-axis corresponds to the boundary of the safety threshold $\delta=5$m.}
    \label{fig:UUVminDists}
    \vspace{-10pt}
\end{figure}

The attractive gradient in \eqref{eqn:fullgrad} is computed using standard gradient techniques. However, each of the $n$ repulsive field gradients requires running reachability analysis, which is generally not computationally feasible at runtime. To get around this issue, we train a neural network to compute the repulsive field gradient of each obstacle.


The NN computes the repulsive field gradient of each obstacle separately. This is due to the compositionality inherent in the linearity of the gradient operator, as shown in \eqref{eqn:fullgrad} and the fact that all obstacles share the same control policy by assumption. The compositionality is a key aspect of our approach, since it means we do not need to know the number of obstacles at training time and the NN based planner can generalize to arbitrary numbers of obstacles.


For each obstacle, the NN takes as input the obstacle information and the time since that information was received and outputs the gradient of the repulsive field given in \eqref{eqn:augmentedRepFieldSingleObst}. Thus, the reachability analysis is baked into the weights of the NN during training, along with the repulsive field calculations. This ensures that our planning framework incorporates the uncertainties from the reachability analysis while running in crowded environments in real time.




We generate the neural network training data by fixing the robot position at $\bm{x}_p$ and iterating over a grid of obstacle states $\bm{o}$, robot headings $\bm{x}_{h}$, and times $t$. For each tuple of these values, we compute the gradient of the repulsive field from \eqref{eqn:augmentedRepFieldSingleObst} for robot location $\bm{x}_p$ assuming obstacle information $\bm{o}$ was received $t$ seconds prior:

\begin{align}
    \nabla \left( \frac{1}{2} K_{r} \left(\frac{1}{\left \lVert \bm{x}_p - \bar{\bm{o}}_{t}(\bm{x}_p) \right \rVert - \delta}\right)^2 \right) \label{eqn:NNOutputForTraining}
\end{align}

So the control output at time $t+t_{0}$ is:
\begin{align}
    \bm{u} &= \nabla \left(\frac{1}{2}K_p (\lvert \lvert \bm{x}_p-\bm{x}_{g}\rvert \rvert^2)\right) + \sum_{i=1}^{n} \text{NN}(\bm{x}_p,\bm{x}_h,\bm{o}_i,t) \label{eq:NNControl}
\end{align}
where NN$(\bm{x}_p,\bm{x}_h,\bm{o}_i,t)$ is the neural network's output when run from robot location $\bm{x}_p$ and heading $\bm{x}_h$ on obstacle information $\bm{o}_i$ from time $t_0$ at the current time $t+t_0$.

\section{Experimental Analysis}
\label{sec:evaluation}

To evaluate our planning framework, we ran our approach on a case study of a UUV crossing a crowded shipping channel in a simulation environment and on unmanned ground vehicles (UGV) in a lab environment. Videos of the experiments can be found in the supplemental material.

\subsection{UUV Case Study}

\begin{figure*}
     \subfigure[Sequence of snapshots for the ground vehicle going to its goal (green square).]{\includegraphics[width=0.78\textwidth]{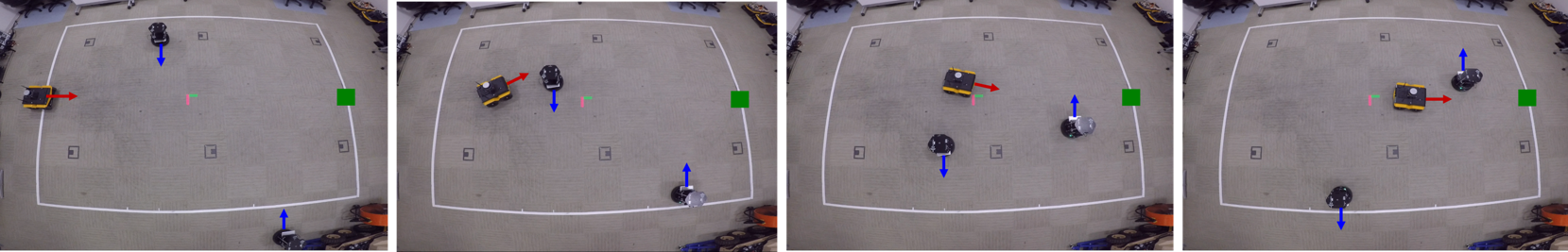}\label{fig:exp_setup_cross}}\hspace{1em}%
    \subfigure[Robots' paths]{\includegraphics[width=0.19\textwidth]{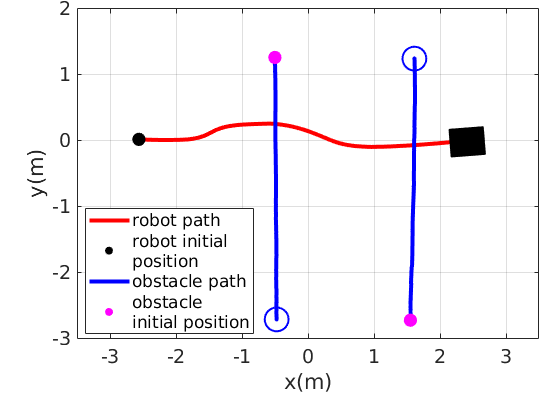}\label{fig:exp_path_cross}}
        \vspace{-5pt}
    \caption{Experiment results with a ground vehicle and two dynamic obstacles moving in the opposite directions.}
    \label{fig:exp_cross}
\end{figure*}

\begin{figure*}
     \subfigure[Sequence of snapshots for the ground vehicle going to its goal (green square).]{\includegraphics[width=0.78\textwidth]{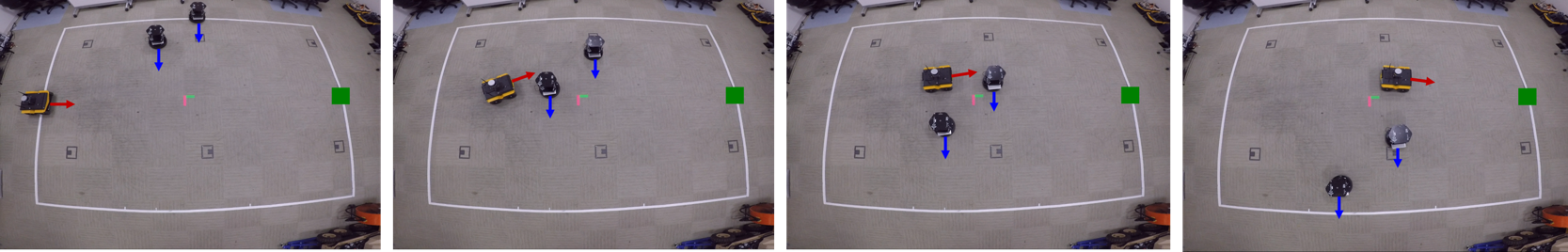}\label{fig:exp_setup_parallel}}\hspace{1em}%
    \subfigure[Robots' paths]{\includegraphics[width=0.19\textwidth]{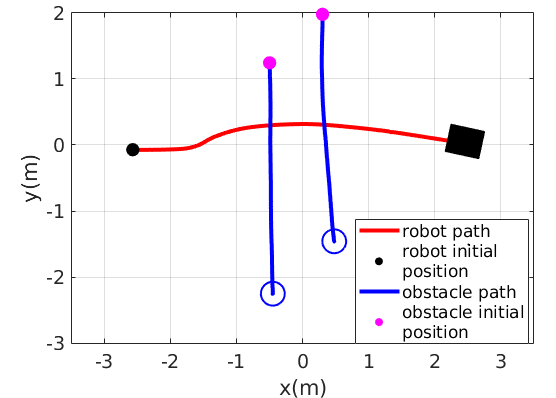}\label{fig:exp_path_parallel}}
        \vspace{-5pt}
    \caption{Experiment results with a ground vehicle and two dynamic obstacles moving in the same direction.}
    \label{fig:exp_parallel}
\end{figure*}

\begin{figure}[h]
    \centering
    \subfigure[Minimum distance for the case shown in Fig.~\ref{fig:exp_cross}.]{\includegraphics[width=0.23\textwidth]{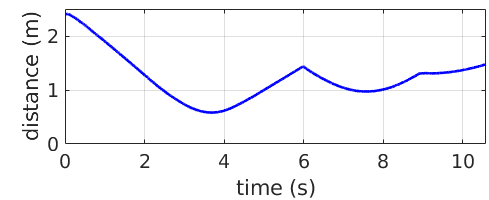}\label{fig:exp_dist_cross}}\hspace{1em}%
    \subfigure[Minimum distance for the case shown in Fig.~\ref{fig:exp_parallel}.]{\includegraphics[width=0.23\textwidth]{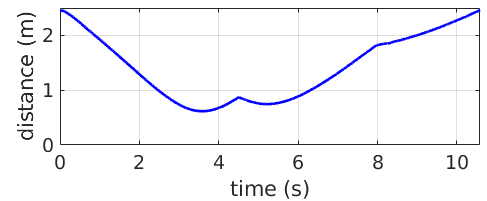}\label{fig:exp_dist_parallel}}
    \caption{Minimum distance between the ground vehicle and two dynamic obstacles over time.}
    \label{fig:experiment}
    \vspace{-10pt}
\end{figure}

For our UUV case study, the scenario we considered is as follows: a UUV needs to cross a shipping channel, in which ships are traveling at a constant $5$m/s with noise bounded by $0.05$m/s, each going opposite directions, similar to how cars would travel on a two-lane road. In addition, all ships have a $0.01$ radian noise bound on their headings. Each ship is $75$m long and $25$m wide and there is a $375$m gap in between each successive ship. The total width of the channel is $230$m. The UUV surfaces around $500$m before the channel to receive the location, heading, and speed of each of the ships. The UUV then dives down to a depth of $45$m and travels to a predefined waypoint directly across the channel. The UUV used for the simulation is $2$m long, $15$cm in radius, and travels at a maximum speed of $2.5$m/s. To ensure that there is always at least a 2 UUV lengths distance to the ships, the safety threshold is set to $\delta=5$m. The simulations were run on a real-time, physics-based ROS Gazebo UUV simulator \cite{Manhes2016}.

The obstacle forward reachable sets were extended out by $T=600$ seconds using Flow* \cite{Chen2013}. We found that this was enough to prevent most collisions while still allowing the UUV to quickly cross the channel. The potential field constants were $K_p=5$ and $K_r=15000$. The NN architecture consisted of 15 fully-connected ReLU layers, each with 16 neurons, along with linear layers on the input and output, and was trained using around 8 million data points\footnote{This number depends on the a priori known size of the environment, the maximum distance between the robot and obstacles, the range of obstacle speeds and the maximum time between intermittent information.}. We trained the NN using the Keras library.



We ran $20$ trials of the UUV crossing the shipping channel, varying the initial positions of the ships traveling in each direction. Our planning framework safely guided the UUV through the channel while avoiding the ship reachable sets (and hence the ships themselves). The closest the UUV came to any ship across all trials was $7.5$m.  One example path can be seen in blue in Fig.~\ref{fig:UUVpath}. We also plotted the minimum distance to the ship reachable sets for each of the 20 trials in Fig.~\ref{fig:UUVminDists}. Each scenario took around $12$ minutes and the UUV got the ship information $190$ seconds into the scenario. So the UUV had to plan using information which was up to $9$ minutes old to complete the scenarios. Finally, the average runtime of each control iteration was $27$ms and the largest runtime was $81$ms, whereas computing the reachable set of each ship $600$ seconds into the future took Flow* an average of $67$ seconds. In fact, this large runtime prohibits the direct application of reachability tools for online safety analysis in dynamic environments.


We also ran the same $20$ trials using the controller from \eqref{eqn:fullgrad} with precomputed reachable sets for each ship, which we refer to as the reach set PF planner. The closest the UUV came to any ship reachable sets was $18.8$m. One example path can be seen in orange in Fig.~\ref{fig:UUVpath}. The minimum distance to the ship reachable sets for each of the 20 trials can be seen in Fig.~\ref{fig:UUVminDists}. This demonstrates both that the control technique proposed in \eqref{eqn:fullgrad} is safe and effective and that our NN reasonably approximates the repulsive field gradients of the ship reachable sets.

Finally, we ran the same 20 trials using the controller from \eqref{eqn:fullgrad} assuming no uncertainty in the ship speeds and headings. One example path can be seen in purple in Fig.~\ref{fig:UUVpath}. The minimum distance to the ship reachable sets for each of the 20 trials can be seen in Fig.~\ref{fig:UUVminDists}. In $11$ of the $20$ trials the UUV came within the allowed $\delta = 5$m of the reachable sets and in one of the trials the UUV actually collided with one of the reachable sets. This collision can be seen in Fig.~\ref{fig:uuvpath3}.

\subsection{UGV Experiments}
To demonstrate the applicability of the approach to different types of vehicles, we conducted unmanned ground vehicle experiments. The experiments were performed on a Clearpath Jackal UGV and two TurtleBot2 UGVs were used as mobile obstacles. In the offline stage, the reachable sets of the dynamic obstacles were extended out by 8 seconds using Flow*\cite{Chen2013} as explained in Section~\ref{sec:reachability} and an NN was trained to compute the desired repulsive fields for each obstacle fast at runtime as presented in Section~\ref{sec:neural_net}. The NN was trained using around 3 million data points and had an architecture of 4 fully-connected ReLU layers, each with 16 neurons, along with linear layers on the input and output. The potential field constants were $K_p=5$ and $K_r=10$.

At runtime, the robot was tasked to reach its goal location at $ [2.5, 0] $m moving with desired speed of $ 0.5 $m/s while avoiding the other two vehicles which moved straight also with a speed of $ 0.5 $m/s. The ground vehicle observed the position and heading of the mobile obstacles only in the beginning of the operation, then lost its connection. We also assumed here that the robot did not have any onboard sensors to detect the obstacles. A Vicon motion capture system was used to detect the state of the ground vehicle and the initial position and heading of the dynamic obstacles.

We tested our approach on two different scenarios. In the first case, the mobile obstacles were moving in opposite directions in between the robot's start and goal locations. Fig.~\ref{fig:exp_setup_cross} demonstrates the behavior of the ground vehicle and mobile obstacles over time. Fig.~\ref{fig:exp_path_cross} shows the actual paths of these vehicles and Fig.~\ref{fig:exp_dist_cross} shows the minimum distance between the ground vehicle and the obstacles over time. For this case study, the minimum distance from any of the obstacles was recorded as $58.26$cm which is larger than the safety threshold $ \delta = 51 $cm (summation of the ground vehicle and obstacle radius), which means that the ground vehicle was able to take necessary actions to avoid both obstacles. The second scenario we tested was with two obstacles moving in the same direction as shown in Fig.~\ref{fig:exp_setup_parallel}. The resultant paths of the vehicles are demonstrated in Fig.~\ref{fig:exp_path_parallel} and the minimum distance to the obstacles over time is shown in Fig.~\ref{fig:exp_dist_parallel}. Similar to the previous case, the ground vehicle successfully avoided collision with both obstacles with the closest distance recorded as 61.60cm.

\section{Conclusion and Future Work}
\label{sec:discussion}

In this work, we have presented a fast path planning framework for operating in dynamic environments with intermittent obstacle information. We combine reachability analysis and artificial potential fields as the backbone of our planning framework and leverage neural networks to ensure our control loop runs in real time. We have applied our approach on a simulation of a UUV crossing a shipping channel and presented experiments with ground vehicles. Thanks to our technique, the robot is able to avoid both collision and getting trapped into dynamical minima. 

In the future, we will consider extensions of the proposed framework to account for additional sources of uncertainty, such as localization uncertainty or model uncertainty of the ego-vehicle. We also plan to consider cases in which some vehicles are cooperative while others are not. Additionally, we will investigate the use of verification tools to provide safety guarantees for the proposed learning-based control framework. Finally, we are planning to investigate the inclusion of this control framework into a confidence monitoring framework by leveraging other sources of measurements on these robot, such as the expected measurements from range sensors during a communication denied mission in a dynamic environment. 

\section*{Acknowledgments} 
This work was supported in part by AFRL and DARPA FA8750-18-C-0090, ARO W911NF-20-1-0080, and ONR N00014-17-1-2012. 

\bibliographystyle{IEEEtran}
\bibliography{sample_base.bib}

\begin{thebibliography}{10}
\providecommand{\url}[1]{#1}
\csname url@samestyle\endcsname
\providecommand{\newblock}{\relax}
\providecommand{\bibinfo}[2]{#2}
\providecommand{\BIBentrySTDinterwordspacing}{\spaceskip=0pt\relax}
\providecommand{\BIBentryALTinterwordstretchfactor}{4}
\providecommand{\BIBentryALTinterwordspacing}{\spaceskip=\fontdimen2\font plus
\BIBentryALTinterwordstretchfactor\fontdimen3\font minus
  \fontdimen4\font\relax}
\providecommand{\BIBforeignlanguage}[2]{{%
\expandafter\ifx\csname l@#1\endcsname\relax
\typeout{** WARNING: IEEEtran.bst: No hyphenation pattern has been}%
\typeout{** loaded for the language `#1'. Using the pattern for}%
\typeout{** the default language instead.}%
\else
\language=\csname l@#1\endcsname
\fi
#2}}
\providecommand{\BIBdecl}{\relax}
\BIBdecl

\bibitem{van2008}
J.~Van~den Berg, M.~Lin, and D.~Manocha, ``Reciprocal velocity obstacles for
  real-time multi-agent navigation,'' in \emph{2008 IEEE International
  Conference on Robotics and Automation}.\hskip 1em plus 0.5em minus
  0.4em\relax IEEE, 2008, pp. 1928--1935.

\bibitem{Fletcher2000}
B.~{Fletcher}, ``Uuv master plan: a vision for navy uuv development,'' in
  \emph{OCEANS 2000 MTS/IEEE Conference and Exhibition. Conference Proceedings
  (Cat. No.00CH37158)}, vol.~1, 2000, pp. 65--71 vol.1.

\bibitem{Zucker2013}
M.~Zucker, N.~Ratliff, A.~D. Dragan, M.~Pivtoraiko, M.~Klingensmith, C.~M.
  Dellin, J.~A. Bagnell, and S.~S. Srinivasa, ``Chomp: Covariant hamiltonian
  optimization for motion planning,'' \emph{The International Journal of
  Robotics Research}, vol.~32, no. 9-10, pp. 1164--1193, 2013.

\bibitem{Byravan2014}
A.~Byravan, B.~Boots, S.~S. Srinivasa, and D.~Fox, ``Space-time functional
  gradient optimization for motion planning,'' in \emph{2014 IEEE International
  Conference on Robotics and Automation (ICRA)}, 2014, pp. 6499--6506.

\bibitem{Men2020}
J.~Men and J.~Requena~Carrión, ``A generalization of the chomp algorithm for
  uav collision-free trajectory generation in unknown dynamic environments,''
  in \emph{2020 IEEE International Symposium on Safety, Security, and Rescue
  Robotics (SSRR)}, 2020, pp. 96--101.

\bibitem{Vannoy2008}
J.~Vannoy and J.~Xiao, ``Real-time adaptive motion planning (ramp) of mobile
  manipulators in dynamic environments with unforeseen changes,'' \emph{IEEE
  Transactions on Robotics}, vol.~24, no.~5, pp. 1199--1212, 2008.

\bibitem{Mcleod2016}
S.~McLeod and J.~Xiao, ``Real-time adaptive non-holonomic motion planning in
  unforeseen dynamic environments,'' in \emph{2016 IEEE/RSJ International
  Conference on Intelligent Robots and Systems (IROS)}, 2016, pp. 4692--4699.

\bibitem{guzzi2013human}
J.~Guzzi, A.~Giusti, L.~M. Gambardella, G.~Theraulaz, and G.~A. Di~Caro,
  ``Human-friendly robot navigation in dynamic environments,'' in \emph{IEEE
  International Conference on Robotics and Automation}, Karlsruhe, Germany, May
  6-10, 2013, pp. 423--430.

\bibitem{aoude2013probabilistically}
G.~S. Aoude, B.~D. Luders, J.~M. Joseph, N.~Roy, and J.~P. How,
  ``Probabilistically safe motion planning to avoid dynamic obstacles with
  uncertain motion patterns,'' \emph{Autonomous Robots}, vol.~35, no.~1, pp.
  51--76, 2013.

\bibitem{renganathan2020towards}
V.~Renganathan, I.~Shames, and T.~H. Summers, ``Towards integrated perception
  and motion planning with distributionally robust risk constraints,''
  \emph{IFAC-PapersOnLine}, vol.~53, no.~2, pp. 15\,530--15\,536, 2020.

\bibitem{Liu2020}
L.~Liu, D.~Dugas, G.~Cesari, R.~Siegwart, and R.~Dubé, ``Robot navigation in
  crowded environments using deep reinforcement learning,'' in \emph{2020
  IEEE/RSJ International Conference on Intelligent Robots and Systems (IROS)},
  2020, pp. 5671--5677.

\bibitem{Vemula2016}
A.~Vemula, K.~Muelling, and J.-O. Oh, ``Path planning in dynamic environments
  with adaptive dimensionality,'' in \emph{SOCS}, 2016.

\bibitem{Khodayi2019}
R.~{Khodayi-mehr}, Y.~{Kantaros}, and M.~M. {Zavlanos}, ``Distributed state
  estimation using intermittently connected robot networks,'' \emph{IEEE
  Transactions on Robotics}, vol.~35, no.~3, pp. 709--724, 2019.

\bibitem{Kantaros2019}
Y.~{Kantaros}, M.~{Guo}, and M.~M. {Zavlanos}, ``Temporal logic task planning
  and intermittent connectivity control of mobile robot networks,'' \emph{IEEE
  Transactions on Automatic Control}, vol.~64, no.~10, pp. 4105--4120, 2019.

\bibitem{aragues2021intermittent}
R.~Aragues, D.~V. Dimarogonas, P.~Guallar, and C.~Sagues, ``Intermittent
  connectivity maintenance with heterogeneous robots,'' \emph{IEEE Transactions
  on Robotics}, vol.~37, pp. 225--245, 2021.

\bibitem{Xu2019}
Z.~{Xu}, F.~M. {Zegers}, B.~{Wu}, W.~{Dixon}, and U.~{Topcu}, ``Controller
  synthesis for multi-agent systems with intermittent communication. a metric
  temporal logic approach,'' in \emph{2019 57th Annual Allerton Conference on
  Communication, Control, and Computing (Allerton)}, 2019, pp. 1015--1022.

\bibitem{Bopardikar2016}
S.~D. Bopardikar, B.~Englot, A.~Speranzon, and J.~van~den Berg, ``Robust belief
  space planning under intermittent sensing via a maximum eigenvalue-based
  bound,'' \emph{The International Journal of Robotics Research}, vol.~35,
  no.~13, pp. 1609--1626, 2016.

\bibitem{Penin2019}
B.~{Penin}, P.~R. {Giordano}, and F.~{Chaumette}, ``Minimum-time trajectory
  planning under intermittent measurements,'' \emph{IEEE Robotics and
  Automation Letters}, vol.~4, no.~1, pp. 153--160, 2019.

\bibitem{Yel2019}
E.~{Yel} and N.~{Bezzo}, ``Fast run-time monitoring, replanning, and recovery
  for safe autonomous system operations,'' in \emph{2019 IEEE/RSJ International
  Conference on Intelligent Robots and Systems (IROS)}, 2019, pp. 1661--1667.

\bibitem{Koohifar2018}
F.~Koohifar, I.~Guvenc, and M.~Sichitiu, ``Autonomous tracking of intermittent
  rf source using a uav swarm,'' \emph{IEEE Access}, vol.~PP, 01 2018.

\bibitem{Seo2019}
H.~{Seo}, D.~{Lee}, C.~Y. {Son}, C.~J. {Tomlin}, and H.~J. {Kim}, ``Robust
  trajectory planning for a multirotor against disturbance based on
  hamilton-jacobi reachability analysis,'' in \emph{2019 IEEE/RSJ International
  Conference on Intelligent Robots and Systems (IROS)}, 2019, pp. 3150--3157.

\bibitem{Ding2011}
J.~{Ding}, E.~{Li}, H.~{Huang}, and C.~J. {Tomlin}, ``Reachability-based
  synthesis of feedback policies for motion planning under bounded
  disturbances,'' in \emph{2011 IEEE International Conference on Robotics and
  Automation}, 2011, pp. 2160--2165.

\bibitem{Malone2017}
N.~{Malone}, H.~T. {Chiang}, K.~{Lesser}, M.~{Oishi}, and L.~{Tapia}, ``Hybrid
  dynamic moving obstacle avoidance using a stochastic reachable set-based
  potential field,'' \emph{IEEE Transactions on Robotics}, vol.~33, no.~5, pp.
  1124--1138, 2017.

\bibitem{Pendleton2017}
S.~D. {Pendleton}, W.~{Liu}, H.~{Andersen}, Y.~H. {Eng}, E.~{Frazzoli},
  D.~{Rus}, and M.~H. {Ang}, ``Numerical approach to reachability-guided
  sampling-based motion planning under differential constraints,'' \emph{IEEE
  Robotics and Automation Letters}, vol.~2, no.~3, pp. 1232--1239, 2017.

\bibitem{Desai2020}
A.~{Desai} and N.~{Michael}, ``Online planning for quadrotor teams in 3-d
  workspaces via reachability analysis on invariant geometric trees,'' in
  \emph{2020 IEEE International Conference on Robotics and Automation (ICRA)},
  2020, pp. 8769--8775.

\bibitem{Akametalu2015}
A.~K. {Akametalu} and C.~J. {Tomlin}, ``Temporal-difference learning for online
  reachability analysis,'' in \emph{2015 European Control Conference (ECC)},
  2015, pp. 2508--2513.

\bibitem{Chiang2017}
H.-T.~L. Chiang, B.~HomChaudhuri, A.~P. Vinod, M.~Oishi, and L.~Tapia,
  ``Dynamic risk tolerance: Motion planning by balancing short-term and
  long-term stochastic dynamic predictions,'' in \emph{2017 IEEE International
  Conference on Robotics and Automation (ICRA)}, 2017, pp. 3762--3769.

\bibitem{Franco2020}
C.~D. {Franco} and N.~{Bezzo}, ``Interpretable run-time monitoring and
  replanning for safe autonomous systems operations,'' \emph{IEEE Robotics and
  Automation Letters}, vol.~5, no.~2, pp. 2427--2434, 2020.

\bibitem{Althoff2014}
M.~{Althoff} and J.~M. {Dolan}, ``Online verification of automated road
  vehicles using reachability analysis,'' \emph{IEEE Transactions on Robotics},
  vol.~30, no.~4, pp. 903--918, 2014.

\bibitem{Herbert2019}
S.~L. {Herbert}, S.~{Bansal}, S.~{Ghosh}, and C.~J. {Tomlin},
  ``Reachability-based safety guarantees using efficient initializations,'' in
  \emph{2019 IEEE 58th Conference on Decision and Control (CDC)}, 2019, pp.
  4810--4816.

\bibitem{Chen2013}
X.~Chen, E.~{\'A}brah{\'a}m, and S.~Sankaranarayanan, ``Flow*: An analyzer for
  non-linear hybrid systems,'' in \emph{Computer Aided Verification},
  N.~Sharygina and H.~Veith, Eds.\hskip 1em plus 0.5em minus 0.4em\relax
  Berlin, Heidelberg: Springer Berlin Heidelberg, 2013, pp. 258--263.

\bibitem{Bansal2017}
S.~{Bansal}, M.~{Chen}, S.~{Herbert}, and C.~J. {Tomlin}, ``Hamilton-jacobi
  reachability: A brief overview and recent advances,'' in \emph{2017 IEEE 56th
  Annual Conference on Decision and Control (CDC)}, 2017, pp. 2242--2253.

\bibitem{Soonho2015}
S.~Kong, S.~Gao, W.~Chen, and E.~Clarke, ``dreach: $\delta$-reachability
  analysis for hybrid systems,'' in \emph{Tools and Algorithms for the
  Construction and Analysis of Systems}, C.~Baier and C.~Tinelli, Eds.\hskip
  1em plus 0.5em minus 0.4em\relax Berlin, Heidelberg: Springer Berlin
  Heidelberg, 2015, pp. 200--205.

\bibitem{Kurzhanskiy2006}
A.~A. {Kurzhanskiy} and P.~{Varaiya}, ``Ellipsoidal toolbox (et),'' in
  \emph{Proceedings of the 45th IEEE Conference on Decision and Control}, 2006,
  pp. 1498--1503.

\bibitem{Vasilopoulos2018}
V.~Vasilopoulos, W.~Vega-Brown, O.~Arslan, N.~Roy, and D.~E. Koditschek,
  ``Sensor-based reactive symbolic planning in partially known environments,''
  in \emph{2018 IEEE International Conference on Robotics and Automation
  (ICRA)}, 2018, pp. 5683--5690.

\bibitem{Manhes2016}
M.~Manh{\~a}es, S.~A. Scherer, M.~Voss, L.~R. Douat, and T.~Rauschenbach, ``Uuv
  simulator: A gazebo-based package for underwater intervention and multi-robot
  simulation,'' \emph{OCEANS 2016 MTS/IEEE Monterey}, pp. 1--8, 2016.

\end{thebibliography}

\end{document}